\documentclass[lettersize,journal]{IEEEtran}
\usepackage{amsmath,amsfonts}
\usepackage{algorithmic}
\usepackage{algorithm}
\usepackage{array}
\usepackage[caption=false,font=normalsize,labelfont=sf,textfont=sf]{subfig}
\usepackage{textcomp}
\usepackage{stfloats}
\usepackage{url}
\usepackage{verbatim}
\usepackage{graphicx}
\usepackage{cite}
\usepackage{multirow} 
\usepackage{booktabs}
\usepackage[table]{xcolor}

\definecolor{cvprblue}{rgb}{0.21,0.49,0.74}

\usepackage[pagebackref,breaklinks,colorlinks]{hyperref}

\usepackage[utf8]{inputenc} 
\usepackage[T1]{fontenc}    
\usepackage[colorlinks]{hyperref} 
\usepackage{amsfonts}       
\usepackage{nicefrac}       
\usepackage{microtype}      
\usepackage{float}
\usepackage{amssymb}
\usepackage{wrapfig}
\usepackage{subfig}
\usepackage{array}
\usepackage{lipsum}
\usepackage{cleveref}

\usepackage{colortbl}
\usepackage{tabulary,overpic}

\definecolor{mygray}{gray}{0.92}
\definecolor{baselinecolor}{gray}{.9}

\usepackage{enumitem}

\newlength\savedwidth

\newlength\savewidth
\newcommand\shline{\noalign{\global\savewidth\arrayrulewidth
                            \global\arrayrulewidth 1pt}
                   \hline
                   \noalign{\global\arrayrulewidth\savewidth}}

\newcommand{\tablestyle}[2]{\setlength{\tabcolsep}{#1}\renewcommand{\arraystretch}{#2}\centering\footnotesize}

\makeatletter
\def\blfootnote{\xdef\@thefnmark{}\@footnotetext}
\makeatother

\newcommand{\eg}[0]{\textit{e.g.},~}
\newcommand{\ie}[0]{\textit{i.e.},~}
\newcommand{\etal}[0]{\textit{et al.}~}

\newcommand{\ourmethod}[0]{\fontfamily{ppl}TrackingMiM}

\begin{document}

\title{TrackingMiM: Efficient Mamba-in-Mamba Serialization for Real-time UAV Object Tracking}

\author{
Bingxi Liu,~\IEEEmembership{Graduated Student Member,~IEEE},
Calvin Chen,~\IEEEmembership{Student Member,~IEEE},
Junhao Li, \\
Guyang Yu,
Haoqian Song,
Xuchen Liu,
Jinqiang Cui,
and Hong Zhang,~\IEEEmembership{Life Fellow,~IEEE}

\thanks{Manuscript received June xx, 2025; revised xx xx, 2025. \textit{(Corresponding authors: Hong Zhang; Jinqiang Cui).}}
\thanks{B. Liu is with the Department of Electronic and Electrical Engineering, Southern University of Science and Technology, Shenzhen 518055, China, and also with Peng Cheng Laboratory, Shenzhen 518066, China (e-mail: liubx@pcl.ac.cn).}
\thanks{C. Chen is with the Department of Applied Mathematics and Theoretical Physics, University of Cambridge, Cambridge, CB3 0WA, England (e-mail: hc666@cam.ac.uk).}
\thanks{J. Li is with the State Grid Corporation of China, Jiangsu 213001, China (e-mail: lijunhao069@gmail.com).}
\thanks{G. Yu is with the East China Institute of Computing Technology, Shanghai 201800, China (e-mail: yuguyangsam@gmail.com.)}
\thanks{H. Song, X. Liu and J. Cui are with Peng Cheng Laboratory, Shenzhen 518066, China (e-mail: {liuxc, cuijq}@pcl.ac.cn).}
\thanks{H. Zhang is with Shenzhen Key
Laboratory of Robotics and Computer Vision, the Department of Electronic and Electrical Engineering, Southern University of Science and Technology, Shenzhen 518055, China (e-mail: hzhang@sustech.edu.cn).}
}


\markboth{Submitted to IEEE Transactions on Automation Science and Engineering,~Vol.~XX, No.~X, July~9999}%
{Hidden \MakeLowercase{\textit{et al.}}: TrackingMiM: Efficient Mamba-in-Mamba Serialization for Real-time  UAV Object Tracking}


\maketitle

\begin{abstract}

The Vision Transformer (ViT) model has long struggled with the challenge of quadratic complexity, a limitation that becomes especially critical in unmanned aerial vehicle (UAV) tracking systems, where data must be processed in real time. In this study, we explore the recently proposed State-Space Model, Mamba, leveraging its computational efficiency and capability for long-sequence modeling to effectively process dense image sequences in tracking tasks. First, we highlight the issue of temporal inconsistency in existing Mamba-based methods, specifically the failure to account for temporal continuity in the Mamba scanning mechanism. Secondly, building upon this insight, we propose~\ourmethod, a Mamba-in-Mamba architecture, a minimal-computation burden model for handling image sequence of tracking problem. In our framework, the mamba scan is performed in a nested way while independently process temporal and spatial coherent patch tokens. While the template frame is encoded as query token and utilized for tracking in every scan.  Extensive experiments conducted on five UAV tracking benchmarks confirm that the proposed~\ourmethod~achieves state-of-the-art precision while offering noticeable higher speed in UAV tracking.

\textit{Note to Practitioners---}This paper addresses the pressing need for real-time processing in UAV vision tracking, where existing high-performance models often suffer from excessive computational demands, limiting their feasibility in dynamic aerial environments. Some approaches attempt to reduce memory and processing time by selectively discarding image information, but they still rely on large models and risk omitting critical visual data. In response, this paper introduces a novel state-space approach that is efficient, accurate, and computationally lightweight, enabling real-time performance even on hardware with as little as 4GB of GPU memory.

\end{abstract}


\section{Introduction}
\label{introduction}

\begin{figure}
    \centering
    \includegraphics[width=1.0\linewidth]{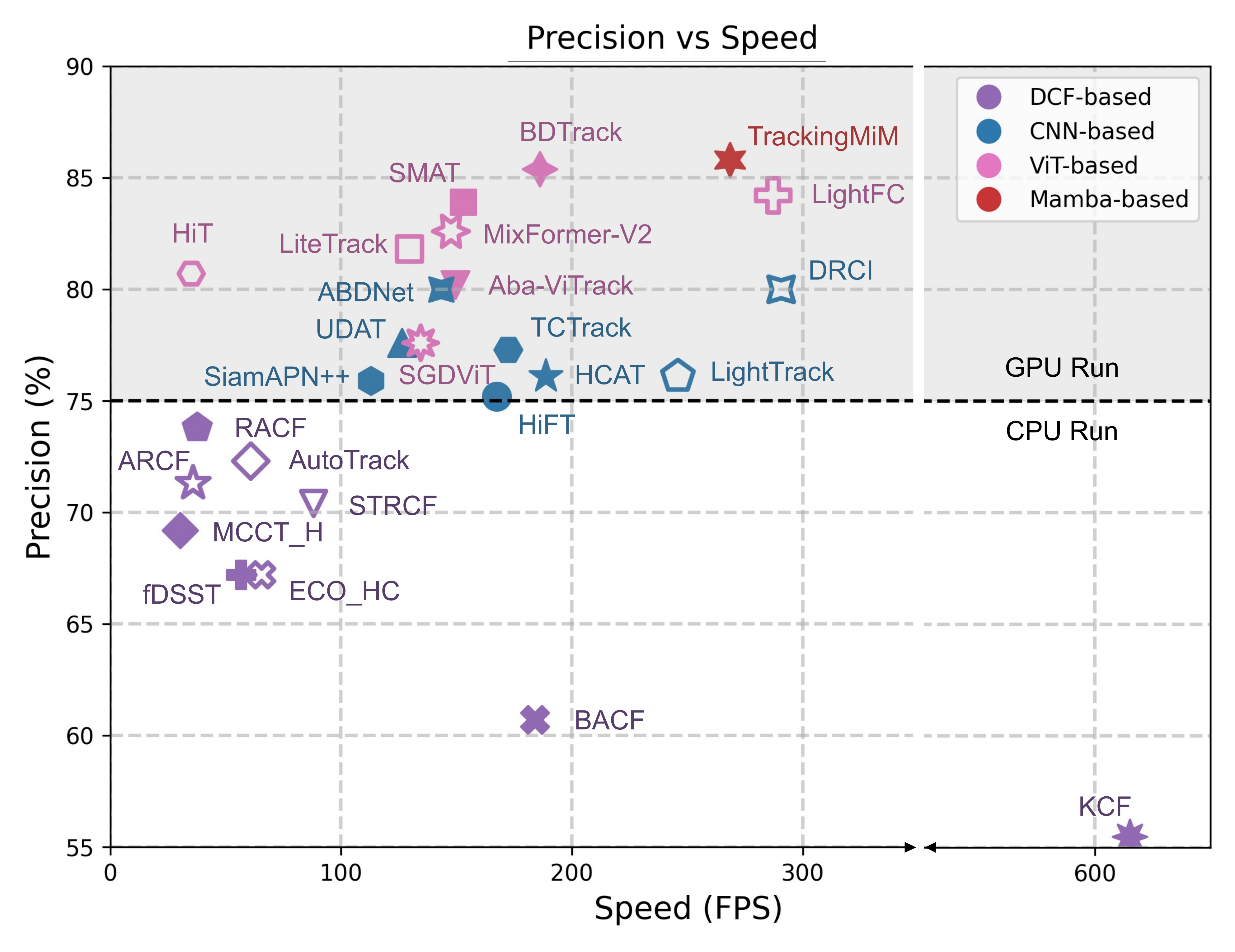}
    \caption{Compared to state-of-the-art UAV tracking algorithms on the UAV123 benchmark, \ourmethod\ achieves a slightly higher precision of 86.6, setting a new record while maintaining efficient performance at 268 FPS.}
    \label{fig:teaser}
\end{figure}

\IEEEPARstart{U}{nmanned} aerial vehicle (UAV) tracking is a critical task that has garnered significant attention due to its essential role in applications such as aerial photography \cite{uav_photo}, security surveillance \cite{uav_surveillance}, and search-and-rescue missions \cite{uav_search}.
UAV tracking entails detecting, predicting, and estimating the position and scale of a target across sequential aerial images captured by high-altitude, mobile cameras~\cite{tase_1, tase_2, uav_vi}. A critical requirement in UAV tracking is real-time performance that ensures continuous and precise monitoring at a minimum frame rate of 30 frames per second (FPS) \cite{li2024learning}. However, achieving real-time tracking is inherently difficult due to several compounding factors. Unlike ground-based systems with access to high-performance computing, UAVs must process tracking data efficiently under stringent power and processing limitations, as they are constrained by the limited computational resources available on onboard hardware \cite{li2023adaptive,wang2023learning}.
In addition to these system-level limitations, rapid motion of either the target or the UAV, extreme viewing angles, motion blur, and low-resolution imagery frequently degrade tracking accuracy. Moreover, occlusions introduce further uncertainty, making reliable tracking even more complex.
To meet the unique demands of UAV applications, an effective tracking system must strike a balance between accuracy, speed, and computational efficiency while operating within the constraints of limited power and processing capacity. 

Current UAV tracking algorithms can be broadly classified into three categories, as illustrated in Fig.~\ref{fig:teaser}: discriminative correlation filters (DCF), convolutional neural networks (CNNs), and vision transformers (ViTs).
Discriminative correlation filter-based trackers~\cite{henriques2015high, kiani2017learning}, which operate in the Fourier domain, are computationally lightweight and efficient. However, these methods often suffer from limited tracking accuracy and robustness, rendering them inadequate for complex UAV tracking scenarios that demand adaptability to dynamic and unpredictable environments.
In contrast, deep-learning-based approaches leverage more sophisticated feature representations to enhance tracking performance. CNN-based trackers~\cite{cao2021hift,cao2021siamapn++,Yan2021LightTrack} excel at learning object features adaptively, achieving high precision. However, the convolutional operations involved are highly computationally intensive, posing challenges for real-time UAV applications. To mitigate this issue, several approaches have been proposed, including lightweight network architectures and branch pruning techniques~\cite{wang2022rank,wu2022fisher,zhong2023fisher} to improve computational efficiency without significantly compromising accuracy.
ViT-based trackers~\cite{ye2022joint,chen2022backbone,kou2023zoomtrack} further advance tracking performance by leveraging self-attention mechanisms, particularly quadratic attention, to model long-range dependencies effectively. While these methods achieve state-of-the-art accuracy, their significantly increased model complexity results in high computational demands, slow inference speeds, and large memory requirements, which hinder their deployment on resource-constrained UAV platforms. 
Notable methods such as SimTrack~\cite{chen2022backbone} and MixFormer~\cite{cui2022mixformer} exemplify the potential of ViT-based tracking in achieving superior accuracy. However, the trade-off between performance and efficiency remains a critical challenge, necessitating further research into practical and computationally efficient UAV tracking solutions.

Mamba \cite{gu2023mamba}, a recently proposed foundational model based on State Space Models (SSMs), has gained widespread attention for its efficiency and strong performance. Unlike traditional transformer-based models, Mamba achieves competitive results in long-sequence modeling tasks while maintaining linear computational complexity, making it a promising alternative in various fields \cite{liu2024vmamba,ye2024p}. However, significant challenges remain in adapting it for tracking tasks, particularly in maintaining tracking continuity and handling occlusions, due to its inherently sequential processing nature, which limits temporal flexibility and complicates the integration of multi-frame contextual cues.

In this work, we explore Mamba as a lightweight model tailored for UAV tracking, aiming to maintain high tracking accuracy while significantly improving efficiency and reducing model size. To this end, we introduce Tracking Mamba-in-Mamba (\ourmethod), a novel framework designed to enhance Mamba's ability to model temporal continuity and spatial detail in UAV tracking tasks.
In tracking applications, frames are serialized as an image sequence, and within the Mamba architecture, images are further patchified into smaller patch sequences ("visual words"~\cite{angeli2008fast}). To fully leverage Mamba's strengths in continuity modeling, we propose a nested Mamba-in-Mamba architecture. The inner Mamba model operates intra-frame, learning fine-grained local features by excavating the relationships within smaller visual words. To further enhance local feature representation, we introduce a window-swing mechanism, which shifts the patch pattern in each block to improve spatial awareness. Then, the visual word features are aggregated and reintegrated into corresponding sequences, ensuring a cohesive and structured representation of the extracted information.
Meanwhile, the outer Mamba model is key to serializable continuous learning, enforcing time consistency across frames. Employing multiple time-scanning schemas better captures long-range dependencies and inter-frame relationships, making it more effective for UAV tracking. 
To address challenges such as occlusion and dynamic object movement, we incorporate query retrieval augmentation tracking, which improves robustness in complex tracking scenarios by refining target re-identification and adaptation over time.

In this article, we introduce the first Mamba-in-Mamba (MiM) architecture specifically designed for UAV object tracking, which we refer to as \ourmethod.  
To summarize, our contributions in this paper are multifaceted, focusing on enhancing tracking performance while simultaneously reducing computational costs:
\begin{enumerate}[leftmargin=*]
    \item Mamba-in-Mamba Architecture:  A nested model design that leverages intra-frame and inter-frame processing for improved feature extraction and temporal continuity in UAV tracking.
    
    \item Time Serialization Scanning: A method of enhancing temporal awareness by systematically arranging and rearranging the scan path of Mamba to optimize the continuity of patches.

    \item Query-Based Retrieval Augmented Tracking:  An adaptive retrieval mechanism that improves target re-identification and robustness, particularly in dynamic and occlusion-heavy tracking scenarios.
    
\end{enumerate}

\section{related work}
\label{related_work}

\subsection{UAV Tracking}

In the field of UAV tracking, modern tracking methods can be broadly categorized into three primary types: DCF-based, CNN-based, and ViT-based approaches. DCF-based trackers are widely utilized in UAV tracking due to their computational efficiency, primarily enabled by the fast Fourier transform (FFT), which facilitates correlation computation in the frequency domain. Their reliance on hand-crafted features ensures low computational overhead, making them particularly suitable for CPU-based implementations \cite{li2021learning,henriques2015high,huang2019learn}. 
However, despite their efficiency, these trackers often struggle with robustness in complex and dynamic environments, as the limited representational capacity of hand-crafted features constrains their ability to adapt to challenging scenarios~\cite{li2020autotrack,kiani2017learning,Liu2022GlobalFP}.
To improve representation capability, numerous studies have explored CNN-based trackers~\cite{cao2021hift,cao2022tctrack}, demonstrating notable advancements in tracking accuracy and robustness for UAV applications. However, their efficiency remains significantly lower than that of DCF-based trackers. While model compression and pruning  techniques~\cite{wang2022rank,wu2022fisher,zhong2023fisher} have been employed to enhance computational efficiency, these approaches often fail to achieve satisfactory tracking precision. Additionally, CNN-based trackers suffer from ineffective template–search correlation, further limiting their performance in UAV tracking scenarios.

Recent advances in visual tracking prioritize unified frameworks with ViTs, presenting numerous representative methods including MixFormer \cite{cui2022mixformer}, AQATrack \cite{xie2024autoregressive}, and EVPTrack \cite{shi2024evptrack}. Xie \etal~\cite{Xie2021LearningTR} introduced a Siamese network that leverages ViT to extract and compare features, facilitating efficient matching. Meanwhile, other approaches favor single-stream architectures that seamlessly integrate processing while reducing model complexity. For instance, TATrack \cite{li2024learning} introduces an efficient one-stream ViT-based tracking framework that seamlessly integrates feature learning and template-search coupling. 
Recent advancements in ViTs have increasingly aimed at enhancing efficiency by optimizing the trade-off between representational capacity and computational cost. This has been achieved through the development of lightweight models, model pruning techniques, and hybrid CNN-ViT architectures~\cite{Zhang2022MiniViTCV,Mao2021TPruneET,Li2022EfficientFormerVT,Chen2021MobileFormerBM}. DynamicViT~\cite{Rao2021DynamicViTEV} enhances token processing efficiency by incorporating control gates that selectively retain relevant tokens. In contrast, A-ViT~\cite{yin2022vit} leverages an adaptive mechanism to eliminate the need for auxiliary halting networks, thereby improving computational efficiency and token prioritization. Similarly, Aba-ViTrack \cite{li2023adaptive} enhances efficiency in real-time UAV tracking using lightweight ViTs and an adaptive background-aware token computation method.

Our work is closely aligned with the Mamba framework, particularly in the context of spatial-temporal modeling for tracking tasks. Notably, several existing studies are highly relevant to our research. MiM-ISTD~\cite{chen2024mim} introduces a nested Mamba architecture for efficient infrared target detection, while Mamba-FETrack~\cite{huang2024mamba} employs the Mamba model for event tracking. In contrast, our approach leverages a specialized Mamba-in-Mamba architecture designed to address spatial-temporal challenges in tracking.

\subsection{State Space Models}

The State Space Model (SSM) was originally developed to characterize dynamic systems~\cite{Gu2021CombiningRC}, leveraging its capability for long-term modeling while addressing constraints related to model capacity and computational efficiency. As an advanced extension of SSM, Mamba has demonstrated exceptional potential for efficiently modeling long sequences, particularly in the visual domain. Recent explorations have led to several innovations based on Mamba. VMamba~\cite{liu2024vmamba} introduces a hierarchical architecture that employs a four-directional scanning strategy to enhance representation learning. VisionMamba~\cite{visionmamba} extends this approach by proposing a bidirectional state-space scanning scheme. Additionally, S4ND~\cite{nguyen2022s4nd} integrates local convolution into the Mamba scanning process. Further advancing this framework, Mamba-ND~\cite{li2024mamba_nd} incorporates multi-dimensional scanning mechanisms within a single Mamba block. Pan-Mamba~\cite{He2024PanMambaEP} employs channel-swapping and cross-modal Mamba to achieve efficient cross-modal information exchange and fusion.

Given the critical role of the scanning schema in the Mamba block for enhancing learning representations, our proposed method, \ourmethod, builds upon prior advancements in optimizing scanning strategies. At its foundation, our method introduces the Mamba-in-Mamba block, a hierarchical architecture that separates spatial and temporal scanning into distinct, independently formulated Mamba blocks. To further enhance efficiency,  we strategically organize temporal and spatial Mamba blocks within precisely designed scanning paradigms, facilitating the seamless integration of spatial and temporal processing.

\subsection{Visual Retrieval Augmentation}
Retrieval augmentation was originally introduced in language generation tasks to enhance parameter efficiency and mitigate hallucination issues. The Retrieval-Augmented Generation (RAG) framework~\cite{lewis2020retrieval} integrates both parametric and non-parametric memory access, enabling more effective generative modeling. More recently, retrieval augmentation has been extensively applied to various computer and robotics vision tasks~\cite{long2022retrieval,xu2021texture,tseng2020retrievegan}. For instance, Long \etal~\cite{long2022retrieval} leverage retrieval-augmented classification to address long-tail visual recognition, while Zhao \etal~\cite{zhao2024retrieval} incorporate retrieval augmentation into few-shot medical image segmentation. RDMs~\cite{blattmann2022retrieval} introduce a method for efficiently storing image databases while conditioning a compact generative model. Kim \etal~\cite{kim2024retrieval} propose retrieval augmentation to the Open-Vocabulary Detection task. RTAGrasp~\cite{dong2024rtagrasp} introduces a retrieval-augmented framework that tasks-oriented grasping constraints from human demonstration videos to novel objects.

Unlike previous works, we are, to the best of our knowledge, the first to apply it to visual tracking challenges.

\begin{figure*}[t]
    \centering
     \includegraphics[width=1\textwidth ]{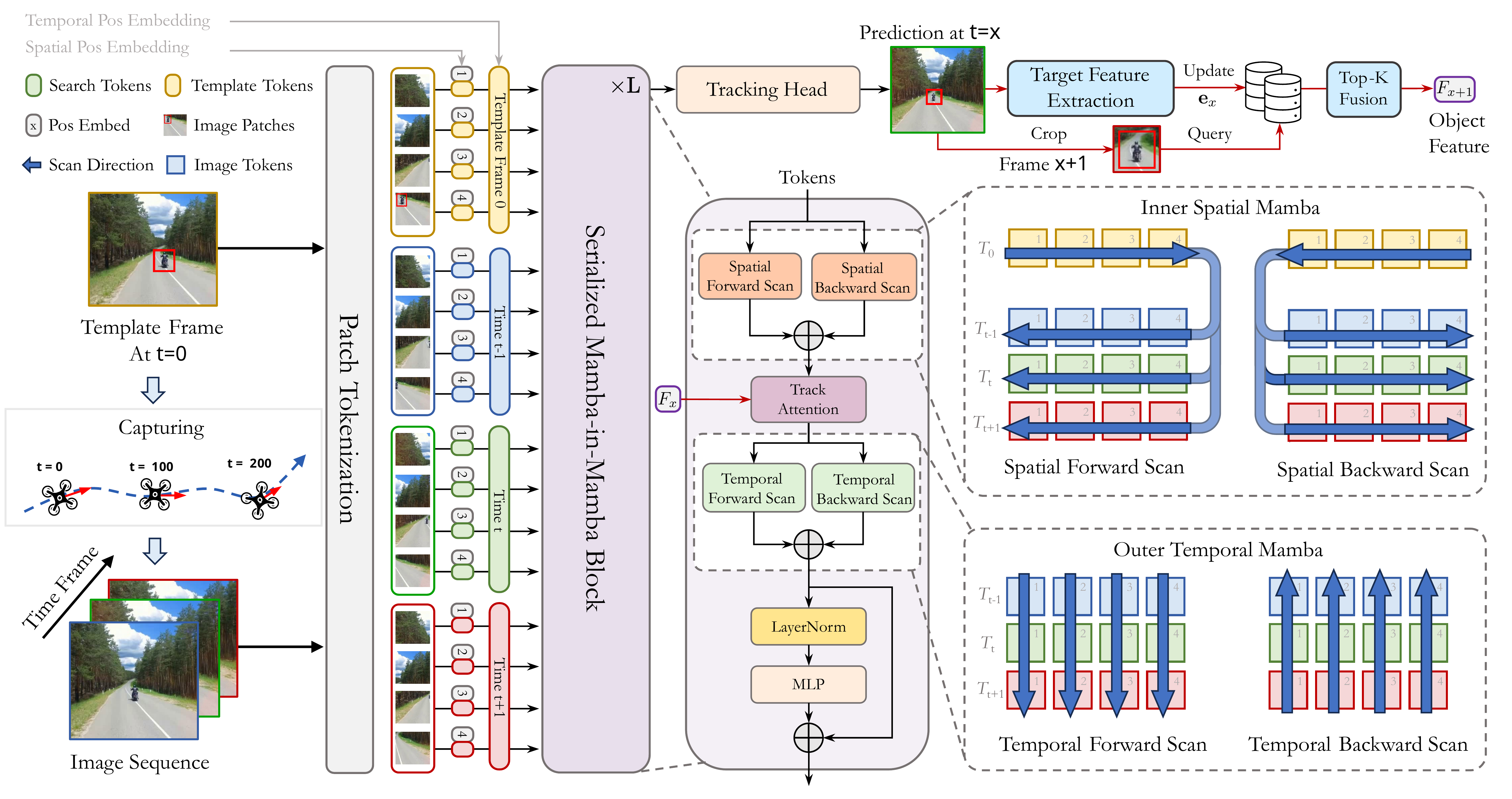}
    \vspace{-0.1cm}
    \caption{\small \textbf{\ourmethod{} Architecture.} The template frame is selected at $t{=}0$ with a bounding box (bbox) input. Subsequent frames are captured and processed within a fixed temporal window. Each frame is first tokenized into patches before being fed into the MiM blocks. In each MiM block, spatial bi-directional scanning is first applied to the template, followed by each frame to integrate template information. A temporal scan then aggregates features across frames at each spatial location. A memory module retrieves the top-$K$ features based on the previous bbox prediction, averages them, and projects the result through an MLP to generate the object feature. Between the spatial and temporal scans, a tracking attention module uses the object feature as a query to attend to the key-value pairs from the spatial outputs.}
    \label{fig:arch}
\end{figure*}


\section{METHOD}
\label{method}

\subsection{Preliminary: SSMs and Mamba}

Mamba serves as a foundational framework based on State Space Models (SSMs), specifically designed for modeling linear time-invariant systems and effectively capturing long-range dependencies. It achieves this by processing an input sequence $ x(t) \in \mathbb{R}^{L} $ through an intermediary hidden state $ h(t) \in \mathbb{R}^{N} $, ultimately generating an output $ y(t) \in \mathbb{R}^{L} $. The behaviour of an SSM is fundamentally dictated by a set of continuous ordinary differential equations (ODEs):

\begin{equation}  
\begin{aligned} 
\label{continuous SSMs}
\dot{h}(t) = {A}h(t) + {B}x(t),    \\  
y(t) = {C}h(t) +  {D}x(t),
\end{aligned}  
\end{equation}

\noindent where $ {A} \in \mathbb{R}^{N \times N} $ denotes the state matrix, $ {B} \in \mathbb{R}^{N \times L} $ represents the input matrix, $ {C} \in \mathbb{R}^{L \times N} $ is the output matrix, and $ {D} \in \mathbb{R}^{L \times L} $ corresponds to the feed-through matrix. The term $ \dot{h}(t) \in \mathbb{R}^{N} $ represents the temporal dynamics of the hidden state.

To apply SSMs in discrete-time settings, the continuous ODEs must first be discretised. Consider a system sampled at discrete time intervals $T = t_{k+1} - t_k $, where $t_k $ and $t_{k+1} $ denote consecutive sampling instants. The transition from the continuous to the discrete domain is achieved using matrix exponentials, yielding the discrete-time state equation:

\[
x(t_{k+1}) = e^{\Delta A}x(t_k) + \left(e^{\Delta A} - I\right)(\Delta A)^{-1}\cdot\Delta  B \cdot u(t_k), 
\]
\noindent where $e^{\Delta A}$  is the matrix exponential represents evolution of the state over the interval $T$, while the term $ \left(e^{\Delta A} - I\right)(\Delta A)^{-1} B$ represents the discrete equivalent of the continuous input effect over the same interval. The parameter $ \Delta $ defines the time scale of discretisation. 

To further refine the discrete representation, the zero-order hold (ZOH) assumption is applied, which leads to a formulation well-suited for numerical computation. Under this assumption, the continuous-time SSM is transformed into its discrete-time equivalent as follows:

$$
 h_k = \mathbf{\bar{A}} h_{k-1} + \mathbf{\bar{B}} x_k,
$$
$$
 y_k = \mathbf{\bar{C}} h_k + \mathbf{\bar{D}} x_k. 
$$
Here, $ \mathbf{\bar{A}} = e^{\Delta A} $, $ \mathbf{\bar{B}} = (e^{\Delta A} - I)(\Delta A)^{-1} \cdot \Delta B $, $ \mathbf{\bar{C}} = C $, and $ \mathbf{\bar{D}} = D $ represent the matrices for the discrete model. 
Mamba harnesses this computational efficiency to enhance sequence modelling in neural networks. Its core computational mechanism involves recursively integrating the previous hidden state $ h_{t-1} $ with the current input $ x_t $, following the formulation:

$$
\begin{aligned}
\overline{\mathbf{K}} & =\left(\mathbf{C} \overline{\mathbf{B}}, \mathbf{C} \overline{\mathbf{A B}}, \ldots, \mathbf{C} \overline{\mathbf{A}}^{m-1} \overline{\mathbf{B}}\right) \\
\mathbf{y} & =x \otimes \overline{\mathbf{K}}
\end{aligned}
$$
Here, $ m $ represents the length of the input sequence $ \mathbf{x} $, $ \otimes $ denotes a convolutional operation, and $ \overline{\mathbf{K}} \in \mathbb{R}^m $ is a convolutional kernel. This structured convolutional transformation enables Mamba to capture long-range dependencies while maintaining computational efficiency effectively. Further details on Mamba can be found in~\cite{Gu2023MambaLS,wang2024SSMSurvey}.

\subsection{Overview: Tracking Mamba-in-Mamba}

Fig.~\ref{fig:arch} delineates the detailed architecture of TrackingMiM (Tracking Mamba-in-Mamba). Our design of TrackingMiM seeks to tackle two interrelated challenges in object tracking: effectively learning tracking representations and optimising the querying of template frames. To systematically address each challenge, we introduce a set of tailored methodologies, each strategically developed to target a specific aspect of the problem. In particular, we propose the Mamba-in-Mamba architecture (\S{\ref{med:mim}}), which incorporates template-first spatial scanning and time serialisation scanning to enhance sequential modelling capabilities. Additionally, we introduce query-based retrieval-augmented tracking (\S{\ref{med:query}}), a new approach designed to optimise query exploitation for more efficient information retrieval and tracking.

\subsection{Mamba-in-Mamba Architecture}
\label{med:mim}

Fig.~\ref{fig:arch} presents an overview of our Mamba-in-Mamba (MiM) architecture. The MiM framework begins with a 3D Patch Tokenisation process, which prepares the input tokens for integration into the Mamba structure. 

\smallskip
\noindent \textbf{Tokenisation.} 
This process is initiated by applying a two-dimensional patchify convolution $\mathcal{P}(\cdot)$ with a kernel of size $ K \times K $ to both the template frame $ \mathbf X^0 \in \mathbb{R}^{C \times H \times W} $ and the input sequence $ \mathbf X^t \in \mathbb{R}^{C  \times H \times W} $ at $T$ timepoints, which frames are indexed sequentially from 1 to $ T $. The convolution operation is applied independently to each frame, which is divided into non-overlapping spatial patches, resulting in $ L $ patches per frame of a fixed size. Each patch is then represented as $ \mathbf{P}^{t,s} \in \mathbb{R}^{C \times K \times K} $, where $ 0 \leq t \leq T $ denotes the temporal frame index, and $ 0 < s \leq L $ corresponds to the spatial patch index. This transformation restructures the image into a sequence of structured patches, optimised for subsequent processing within the Mamba architecture. In our tasks, the parameter $ C=3 $ corresponds to an RGB-channel image, while the kernel size was set to $ K=36 $. 
$$
[\mathbf{P}^{t,p}] = \mathcal{P}(\mathbf X^{t})
$$

The TrackingMiM encoder processes a sequence of input tokens represented by:
$$
\mathbf{P}=\left[\mathbf{P}^{0,p}, \mathbf{P}^{t,p}\right]+\mathbf{e}_s+\mathbf{e}_t.
$$
In this formulation, $\mathbf{P}\in\mathbb{R}^{T+1,L}$ is the new patch token vector. the term $ \mathbf{e}_s \in \mathbb{R}^{L \times C \times K \times K} $ represents a learnable spatial position embedding, which encodes the positional dependencies of individual patches within each frame. Additionally, to preserve temporal coherence and enhance the sensitivity of Mamba to token order, an auxiliary temporal position embedding, $ \mathbf{e}_t \in \mathbb{R}^{(T+1)\times C \times K \times K} $, is incorporated.

\smallskip
\noindent \textbf{Mamba-in-Mamba Block.} 
Following the restructuring of spatial and temporal patch indices, we introduce the Mamba-in-Mamba block, a hierarchical nested framework designed to refine the modelling of spatiotemporal tracking features. This architecture unfolds in two sequential stages, beginning with a spatial Mamba block that initially processes patches within individual frames before extending its operation across the spatial domain. By first capturing local spatial interactions within a given frame, this stage establishes a foundation for more structured feature extraction.

Building upon this, the second stage incorporates a temporal Mamba block, facilitating the propagation of extracted features across $T$ frames. This temporal processing follows a bidirectional scanning strategy, enabling a more comprehensive encoding of long-range dependencies while preserving temporal coherence. Notably, all scanning operations, whether in the spatial or temporal domain, consistently originate from the reference frame, ensuring a well-structured and coherent reference information flow throughout the sequence.


\smallskip

\noindent \textbf{Template-first Spatial Scanning.}  
To maintain spatial continuity and reduce object fragmentation caused by rigid patch partitioning, we propose template-first spatial scanning—an adaptive method that dynamically adjusts the partitioning strategy across layers. 
Given patches $\mathbf{P}^{{t},p}$ at timepoint ${t}$, $\Omega_i$ denotes the scanning strategy at layer $i$. The template feature is computed as:

\begin{equation}
    [\mathbf{P}^0_i] = \text{scan}(\mathbf{P}^{0,p}_{i-1}, \overline{\mathbf{K}}^p_i , \Omega_i)\Big|_{p \in \mathcal{I}},
\end{equation}

\noindent where $\mathcal{I}$ denotes the index set of spatial patches. To query the template effectively, the scanning operation for each frame in the image sequence is defined as:

\begin{equation}
\begin{aligned}
[\mathbf{P}^t_i] = \text{scan}\left(\sum_{j=1}^{L} s_j \cdot \mathbf{P}^{0,j}_i + \mathbf{P}^{t,p}_{i-1}, \overline{\mathbf{K}}^{t \cdot L + p}_i, \Omega_i\right)\Big|_{p \in \mathcal{I}},
\end{aligned}
\end{equation}

\noindent  where $ s_j $ is a learned spatial attention score obtained through an attention module over the template query, which is then summarized into an overall template token ($\sum_{j=1}^{L} s_j \cdot {\mathbf{P}^{0,j}_i}$).

\smallskip
\noindent \textbf{Time Serialisation Scanning.} 
In traditional scanning methods, spatial scanning was performed without an explicit time-dependent component, assuming a quasi-static or steady-state system. However, in dynamic environments where the observed system evolves over time, neglecting temporal variations leads to incomplete or inaccurate reconstructions. To address this limitation, the Mamba framework incorporates a dedicated temporal scanning mechanism, ensuring that dynamic variations are explicitly captured.

Formally, a purely spatial scan captures a static snapshot $\mathbf{P}^t$, which is insufficient when $\frac{d\mathbf{P}}{dt} \neq 0$, where  $d\mathbf{P}^t = \mathbf{P}^{t}_i - \mathbf{P}^{t}_{i-1}$ represents patch residual. The temporal scanning approach introduces a discrete sampling over time:
\begin{equation}
\mathbf{P}^{t}_i = \mathbf{P}^{t}_i + \sum_{i=0}^{t} \Delta t\cdot   \mathbf{K}^{t}_i \cdot \frac{d\mathbf{P}^t}{dt}\Big| t \in \mathcal{T},
\end{equation}
 $\Delta t$ is a fixed frame time difference term that defines the temporal resolution.  $\mathcal{T}$ represents the index set of  the temporal patches at the same spatial location. This gives time resolution properly reconstructed based on:  $\left|\frac{d\mathbf{P}^t}{dt}\right|$ given any $\Delta t$, mitigating errors due to time evolution.

\subsection{Vision-based Retrieval Augmented Tracking}
\label{med:query}

We introduce \textit{Retrieval-Augmented Tracking} (RAT), a module inserted between the spatial and temporal scans in each MiM block. RAT retrieves historical tracking features to guide object localization. A lightweight Mamba network serves as the feature encoder: the input is first cropped based on the bounding box, with the longer side resized to 64x64, then input to the feature encoder and get the feature $\mathbf e$.

\smallskip
\noindent \textbf{Construct and Update.}
The memory corpus is maintained as a set of feature embeddings $\mathbf{C} = \{\mathbf{e}_1, \ldots, \mathbf{e}_n\}$, and is updated based on cosine similarity to suppress redundancy. A new feature $\mathbf{e}_q$ is added only if it is sufficiently dissimilar from existing entries. Specifically, the update is performed when its maximum similarity to the corpus falls below a threshold $\tau = 0.8$:

\begin{equation}
    \max_{\mathbf{e}_i \in \mathbf{C}} \frac{\mathbf{e}_q \cdot \mathbf{e}_i}{\|\mathbf{e}_q\| \, \|\mathbf{e}_i\|} < \tau.
\end{equation}

The corpus is dynamically updated at inference time to adapt to evolving target appearances.

\newcommand{\CR}{\mathbf{S}_k}

\smallskip
\noindent \textbf{Retrieval.}
Given the current central frame, we use the bounding box from the previous frame, enlarged by a factor of $1.1$, to ensure the object is fully covered. The cropped region is encoded into a query feature $\mathbf{e}_q$. The retriever is defined as a function $\mathcal{R} : (\mathbf{e}_q, \mathbf{C}) \rightarrow \mathbf{S}_k \subset \mathbf{C}$, which takes the query and the corpus as inputs and returns a set of top-$K$ feature candidates ($\mathbf{S}_k$) based on cosine similarity. Formally, the selected retrieval set is:
\begin{equation}
  \mathbf{S}_k = \arg\max_{\substack{\mathbf{S} \subseteq \mathbf{C}, |\mathbf{S}| = k}} \sum_{\mathbf{e}_i \in \mathbf{S}}  \frac{\mathbf{e}_q\cdot \mathbf{e}_i}{\|\mathbf{e}_q\| \|\mathbf{e}_i\|}.
\end{equation}
This retrieval ensures that only the most relevant historical features are selected to guide the current tracking step.


\smallskip
\noindent \textbf{Track Attention.}
We introduce a cross-attention mechanism, denoted as $\mathcal{A} : (\mathbf{e}_q, \mathbf{S}_k) \rightarrow \mathbf{e}_a$, to enrich the query representation by leveraging the retrieved historical features $\mathbf{S}_k$. Specifically, we average the top-$K$ retrieved features to form a fused representation, which is then projected via an MLP to obtain the augmented feature $\mathbf{e}_a$. This serves as the query signal for the tracking attention module.

The tracking attention operates as a cross-attention layer between the spatial and temporal Mamba blocks at every MiM layer. It enhances target awareness by conditioning on the query feature $\mathbf{e}_a$ and attending to the latent representation $\mathbf{h}$ from the spatial Mamba output. The key components are computed as:

\begin{equation} 
    \mathbf{Q}_{a} = \mathbf{W}_{\!\scriptscriptstyle Q} \mathbf{e}_a, \quad 
    \mathbf{K} = \mathbf{W}_{\!\scriptscriptstyle K} \mathbf{h}, \quad 
    \mathbf{V} = \mathbf{W}_{\!\scriptscriptstyle V} \mathbf{h}.
\end{equation}

where $\mathbf{W}_{\!\scriptscriptstyle Q}, \mathbf{W}{\!\scriptscriptstyle K}, \mathbf{W}_{\!\scriptscriptstyle V}$ are learnable projection matrices. The output of the tracking attention is given by the scaled dot-product attention:

\begin{equation}
\hat{\mathbf{h}} = \text{softmax}\left( \frac{\mathbf{Q}_a \mathbf{K}^\top}{\sqrt{d_k}} \right) \mathbf{V},
\end{equation}

\noindent where $\hat{\mathbf{h}}$ denotes the updated latent representation from the model, and $d_k$ is the dimensionality of the key vectors. This enables precise and robust localization by dynamically aligning the query with context-aware representations, effectively acting as object cues across the network.

\begin{figure*}
    \centering
    \includegraphics[width=1.0\linewidth]{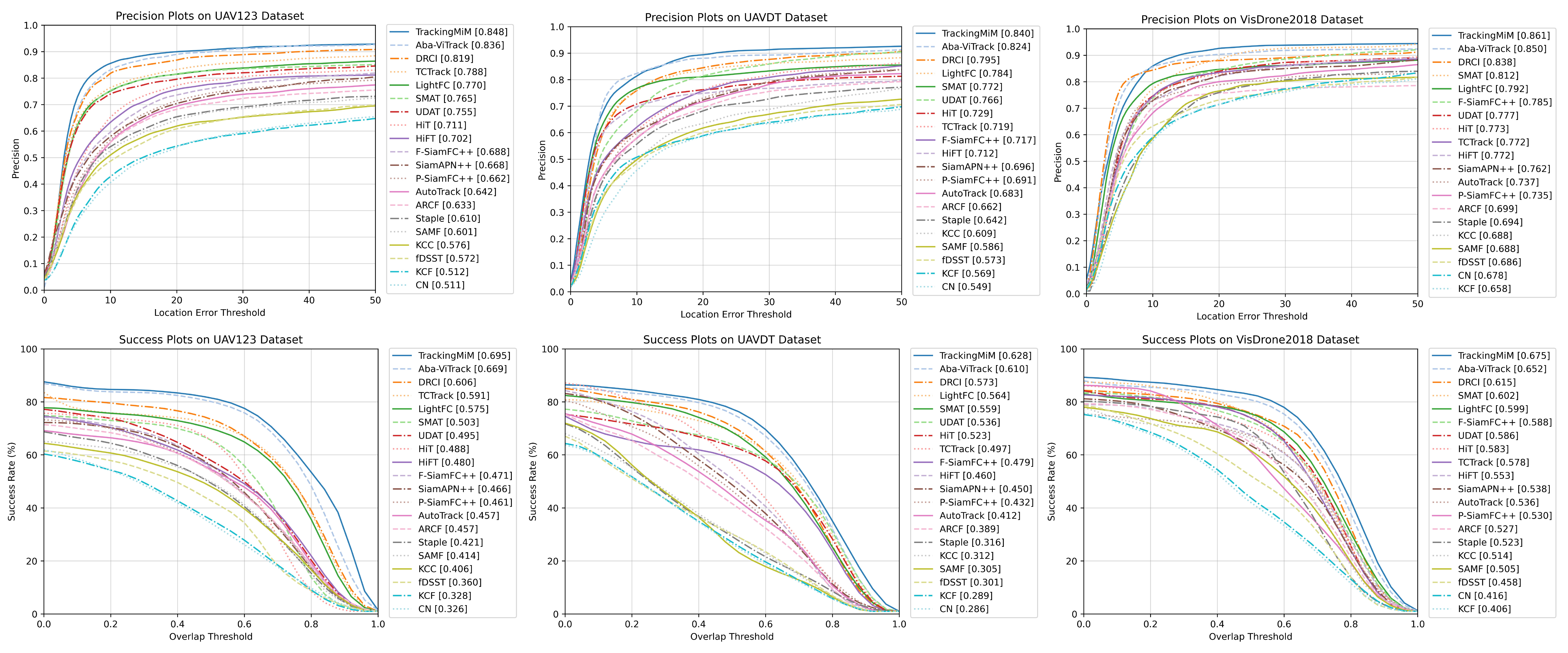}
    \caption{Precision and Success rates versus overlap thresholds on three datasets: UAV123, UAVDT, and VisDrone2018. AUC-based rankings are shown on the right side of each plot.}
    
    \label{fig:result_plot}
\end{figure*}

\newcommand{\first}[1]{\textcolor[HTML]{FE0000}{\textbf{#1}}}
\newcommand{\second}[1]{\textcolor[HTML]{3531FF}{\textbf{#1}}}
\newcommand{\third}[1]{\textcolor[HTML]{1f821c}{\textbf{#1}}}

\begin{table*}[t]
\scriptsize
\centering
\setlength\tabcolsep{5pt} 
\caption{
Comparison of representative trackers in terms of precision (Prec.), success rate (Succ.), and speed (FPS) on five UAV benchmarks: DTB70, UAVDT, VisDrone2018, UAV123, and UAV123@10fps. Prec. and Succ. are shown as percentages (symbol omitted). \first{First}, \second{second}, and \third{third} best results are color-highlighted.  W/o Temporal denotes the use of a Mamba block with only spatial scanning and injected tracking attention. W/o Retrieval removes the tracking attention, using only the MiM block for spatiotemporal modeling. * indicates results reported from the paper due to unavailable official code.}

\label{tab:main}

\begin{tabular}{c|c|c|cc|cc|cc|cc|cc|cc|cc}
\toprule[1pt] 
\multicolumn{2}{c|}{} &  & \multicolumn{2}{c|}{DTB70}& \multicolumn{2}{c|}{UAVDT}& \multicolumn{2}{c|}{VisDrone} & \multicolumn{2}{c|}{UAV123}& \multicolumn{2}{c|}{UAV123@10fps}& \multicolumn{2}{c|}{Avg.}& \multicolumn{2}{c}{Avg. FPS} \\
\multicolumn{2}{c|}{\multirow{-2}{*}{Method}} & \multirow{-2}{*}{Source} & Prec.& Succ.& Prec.& Succ.& Prec.& Succ.& Prec.& Succ.& Prec.& Succ.& Prec.& Succ.& GPU & CPU  \\  \midrule
  & KCF\cite{henriques2015high}  & TPAMI'15  & 49.6 & 31.7 & 61.3 & 30.7 & 70.0 & 43.1 & 54.2 & 35.8 & 42.4 & 28.7 & 55.5 & 34.0 & - & \first{615.0} \\
  & BACF\cite{kiani2017learning} & ICCV'17& 58.8 & 40.8 & 70.8 & 43.2 & 78.8 & 57.5 & 68.6 & 45.9 & 59.1 & 41.1 & 67.2 & 45.7 & - & 56.6 \\
  & fDSST\cite{danelljan2017discriminative}& TPAMI'17  & 54.4 & 37.1 & 69.0 & 38.0 & 69.3 & 51.0 & 58.6 & 40.3 & 52.1 & 39.6 & 60.7 & 41.2 & -& \second{183.8} \\
  & ECO\_HC\cite{danelljan2017eco} & CVPR'17& 65.8 & 44.3 & 69.6 & 41.5 & 81.8 & 58.4 & 70.7 & 52.3 & 64.0 & 46.5 & 70.4 & 48.6 & - & {88.1}  \\
  & MCCT\_H\cite{wang2018multi} & CVPR'18& 60.9 & 41.9 & 66.5 & 41.8 & 79.6 & 59.7 & 67.5 & 45.4 & 61.6 & 43.1 & 67.2 & 46.4 & - & 65.6 \\
  & STRCF\cite{li2018learning}& CVPR'18& 67.0 & 45.3 & 65.2 & 41.4 & 79.7 & 56.4 & 70.4 & 48.4 & 63.8 & 46.8 & 69.2 & 47.7 & - & 30.2 \\
  & ARCF\cite{huang2019learn} & ICCV'19& 71.0 & 48.1 & 71.4 & 46.3 & 81.1 & 57.8 & 66.6 & 48.4 & 66.5 & 48.1 & 71.3 & 49.7  & - & 35.9 \\
  & AutoTrack\cite{li2020autotrack}  & CVPR'20& 71.7 & 50.4 & 72.9 & 46.8 & 78.6 & 58.6 & 68.7 & 47.2 & 69.5 & 50.2 & 72.3 & 50.6 & - & 60.9 \\
\multirow{-9}{*}{\rotatebox{90}{DCF-based}}  & RACF\cite{li2022learning} & PR'22  & 71.2 & 52.0 & 74.3 & 51.8 & 85.7 & {63.4} & 69.1 & 48.9 & 68.6 & 47.9 & 73.8 & 52.8 & - & 37.7 \\ \midrule
  & C2FT\cite{tase_1}* & TASE'19 & - & - & - & - & - & - & 68.7 & 48.5 & - & - & - & - & - & -\\
  & HiFT\cite{cao2021hift} & ICCV'21& 80.7 & 63.0 & 67.0 & 47.0 & 73.5 & 54.2 & 80.0 & 60.9 & 74.7 & 59.6 & 75.2 & 56.9 & {167.5} & -\\
  & SiamAPN++\cite{cao2021siamapn++}  & IROS'21& 78.7 & 60.7 & 79.3 & 55.5 & 74.3 & 56.1 & 78.2 & 58.2 & 76.2 & 61.5 & 77.3 & 58.4 & {172.4} & - \\
  & LightTrack\cite{Yan2021LightTrack} & CVPR'21& 75.7 & 59.8 & 81.0 & \first{64.4} & 74.3 & 58.4 & 80.7 & 64.4 & 76.2 & 60.1 & 77.6 & 61.4 & 126.3 & - \\
  & HCAT\cite{chen2022efficient} & ECCV'22& 83.1 & \third{65.8} & 75.3 & 55.8 & 76.7 & 57.3 & 83.3 & 66.0 & 82.5 & 65.8 & 80.2 & 62.1 & 149.5 & - \\
  & TCTrack\cite{cao2022tctrack} & CVPR'22& 81.1 & 64.9 & 74.5 & 55.1 & 81.2 & 62.2 & 82.4 & 61.3 & 80.6 & 61.7 & 80.0 & 61.0 & 143.3 & - \\
  & UDAT\cite{ye2022unsupervised} & CVPR'22& 83.2 & 64.5 & 82.2 & 59.0 & 81.4 & 63.5 & 76.4 & 58.5 & 80.4 & 61.7 & 80.7 & 61.4 & 35.1 & - \\
  & ABDNet\cite{zuo2023adversarial} & RAL'23 & 77.8 & 59.1 & 76.4 & 58.6 & 77.0 & 56.9 & 79.4 & 63.6 & 77.6 & 60.2 & 77.6 & 59.7 & 134.5 & - \\
\multirow{-9}{*}{\rotatebox{90}{CNN-based}} & DRCI\cite{zeng2023towards} & ICME'23& 81.3 & 62.1 & 83.2 & 60.4 & \third{85.9} & 
63.2 & 76.3 & 61.0 & 73.4 & 54.8 & 80.0 & 60.3 & \third{290.6} & 64.1  \\ \midrule

  & Aba-ViTrack\cite{li2023adaptive}& ICCV'23& \second{85.9} & \second{66.2} & 83.3 & 60.3 & \second{86.3} & 64.3 & \third{86.6} & \third{66.9} & \third{85.0} & \third{66.0} & \second{85.4} & 64.9 & 186.0 & 52.7  \\
  & HiT~\cite{Kang2023ExploringLH}  & ICCV'23& 76.6 & 60.2 & 62.7 & 47.6 & 75.0 & 61.9 & 82.5 & \second{67.1} & 83.9 & \third{66.0} & 76.1 & 60.6 & 245.9 & 59.4 \\
  & LiteTrack~\cite{wei2023litetrack}  & arXiv'23  & 83.2 & 65.1 & 82.2 & 59.9 & 80.4 & 61.2 & 84.1 & \third{66.9} & 83.2 & 64.9 & 82.6 & 63.6 & 147.6 & - \\
   & SGDViT\cite{yao2023sgdvit}  & ICRA'23& 78.4 & 62.7 & 66.1 & 50.0 & 72.2 & 54.3 & 76.1 & 59.5 & \first{86.9} & \first{67.5} & 75.9 & 58.8 & 112.8 & - \\
& MixFormer-V2~\cite{cui2023mixformerv2} & NIPS'23 & 77.3 & 59.6 & 62.1 & 44.5 & 73.3 & 53.4 & 84.1 & 67.7 & 83.7 & 65.4 & 76.1 & 58.1 & 188.7 &  39.2
\\
  & SMAT\cite{gopal2024separable} & WACV'24& 82.6 & 65.4 & 80.4 & 60.2 & 83.1 & 63.3 & 81.4 & 63.8 & 81.5 & 64.1 & 81.8 & 63.4 & 129.6 & -\\
  & LightFC\cite{Li2024LightweightFS} & KBS'24 & 82.8 & 63.4 & \third{84.1} & 60.2 & 82.1 & \second{65.0} & \first{87.6} & 64.8 & 82.8 & 63.3 & 83.9 & 63.3 & 153.0 & - \\
  
\multirow{-8}{*}{\rotatebox{90}{VIT-based}}  &   
BDTrack~\cite{wu2024learningmotionblurrobust}* & arXiv'24 & 83.5 & 64.1 & \third{84.1} & \third{61.0} & 85.2 & 64.3 & 84.8 & 66.7 & 83.5& 65.9 & 84.2 & 64.4 & 287.2* & 63.9*  \\ 

\midrule
\rowcolor{gray!15}
& & w/o Both     & 81.8 & 63.7 & 81.9 & 59.1 & 82.4 & 62.5 & 81.6 & 64.0 & 81.7 & 64.6 & 81.8 & 63.5 & \first{312.9} & \third{129.3}   \\
\rowcolor{gray!15}
& & w/o Retrieval & 84.3 & 65.0 & \second{84.2} & 60.2 & 83.1 & 63.8 & 84.1 & 65.8 & 84.4 & 65.3 & 83.8 & \third{64.5} & 281.6   & 107.5   \\
\rowcolor{gray!15}
& &  w/o Temporal  & \third{84.9} & 65.7 & \third{84.1} & 60.7 & 85.5 & \third{64.4} & 85.9 & 66.5 & 84.8 & 65.8 & \third{84.7} & \second{64.6} & \second{297.1}   & 109.7   \\
\rowcolor{gray!15}
\multirow{-4}{*}{\cellcolor{gray!15}\rotatebox{90}{{Mamba}}} 
& \multirow{-4}{*}{\cellcolor{gray!15}{\ourmethod}} 
& Proposed     & \first{86.7} & \first{67.8} & \first{85.0} & \second{62.4} & \first{86.8} & \first{66.2} & \second{87.1} & \first{68.0} & \second{86.1} & \second{67.1} & \first{86.3} & \first{66.1} & 268.3   & 97.2   \\

\bottomrule[1pt]


\end{tabular}
\end{table*}


\section{experiments}
\label{experiments}

In this section, we first introduce the implementation details and evaluation protocol in \S{~\ref{imple}} and \ref{eval}, respectively. Then, a comprehensive comparison with state-of-the-art methods is presented and quantitatively analyzed in \S{\ref{quantitative}}. In addition, qualitative visualizations of some representative methods are provided in \S{\ref{qualitative}}. Moreover, extensive ablation studies are conducted in \S{\ref{ablation}}, including component-wise removal within our method, as well as insertion within other methods. Finally, \S{\ref{interpretable}} explores the interpretability of our method through feature activation maps.

\subsection{Implementation Details}
\label{imple}

We build our method upon the Mamba architecture, adopting a medium-scale configuration to balance speed and accuracy. Specifically, we employ 24 blocks with a hidden state dimension of 384. The input is tokenized with a temporal stride of 2  with temporal length 8 and spatial token resolution of  16×16. The prediction head is randomly initialized and follows the tracking head design of Aba-ViTrack \cite{li2023adaptive}, with both the search frame and template fixed at 256×256. 

Training is performed using the AdamW optimizer and train for 500 epoch with an initial learning rate of 3e-4, scheduled via 1 epoch of linear warm-up followed by cosine decay. We use a batch size of 8, and apply data augmentations including random bounding box shift, and scale. The training set configuration is aligned with the protocol established in Aba-ViTrack \cite{li2023adaptive}. 

 All experiments are conducted on an NVIDIA GeForce RTX 3090 Ti (24 GB) GPU, paired with an Intel Core i9-13900K (5.8 GHz) CPU. 

\subsection{Evaluation details}
\label{eval}

We evaluate our method on five widely adopted UAV tracking benchmarks: UAV123 \cite{mueller2016benchmark}, UAV123@10fps \cite{mueller2016benchmark}, VisDrone2018 \cite{wen2018visdrone}, UAVDT \cite{du2018unmanned}, and DTB70 \cite{li2017visual}. To ensure fair and comprehensive comparisons, we benchmark against 25 state-of-the-art lightweight trackers, spanning three representative categories: DCF-based, CNN-based, and ViT-based methods (see Tab.~\ref{tab:main}).

\begin{figure*}[t]
    \centering
    \includegraphics[width=1\linewidth]{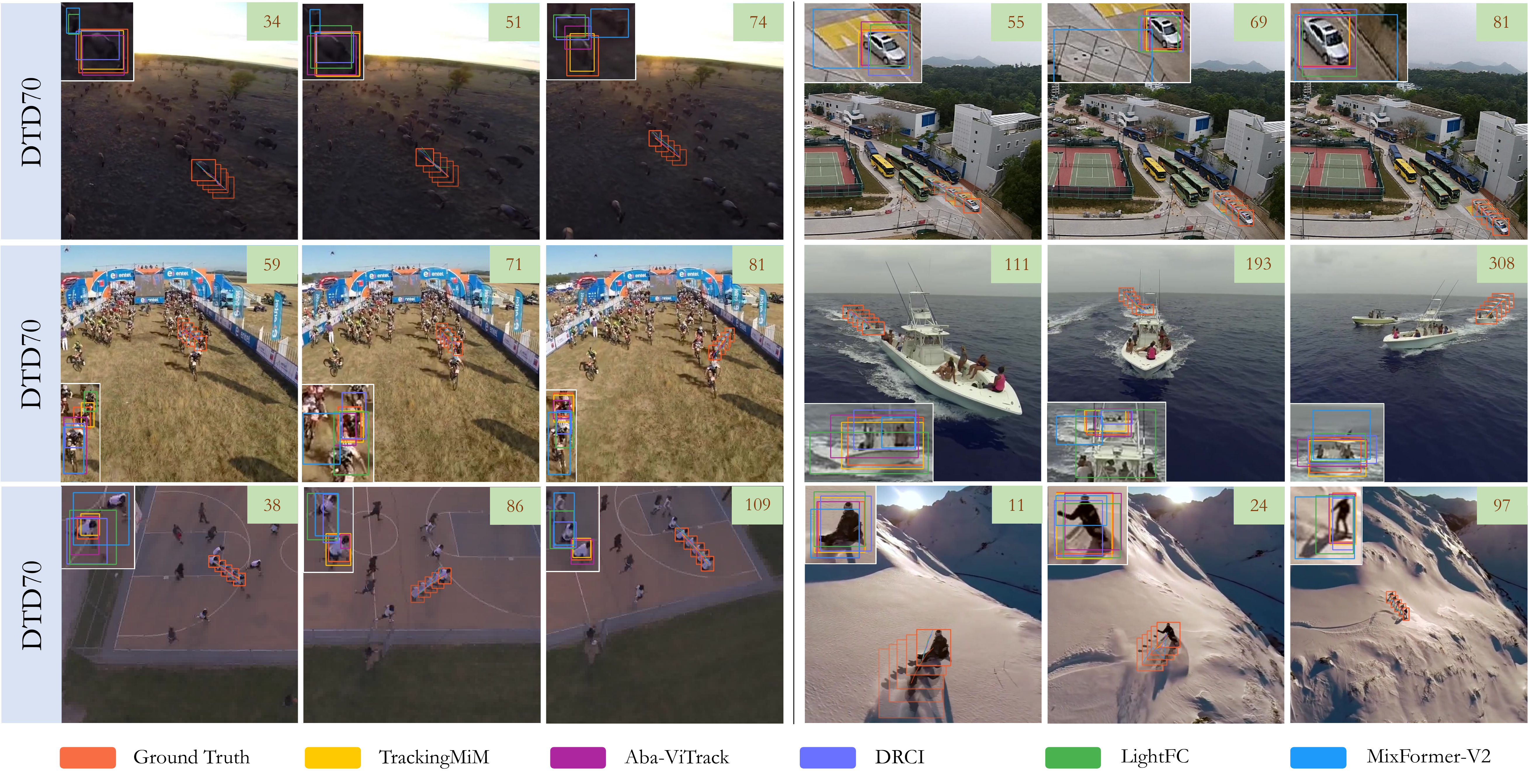}
    \caption{Qualitative evaluation on 6 video sequences from DTB70 (\ie \textit{Animal1}, \textit{Car5}, \textit{MountainBike1}, \textit{Yacht4}, \textit{StreetBasketball2}, and \textit{SnowBoarding4}).}
    \label{fig:visualization_dtd}
\end{figure*}

\begin{figure*}[t]
    \centering
    \includegraphics[width=1\linewidth]{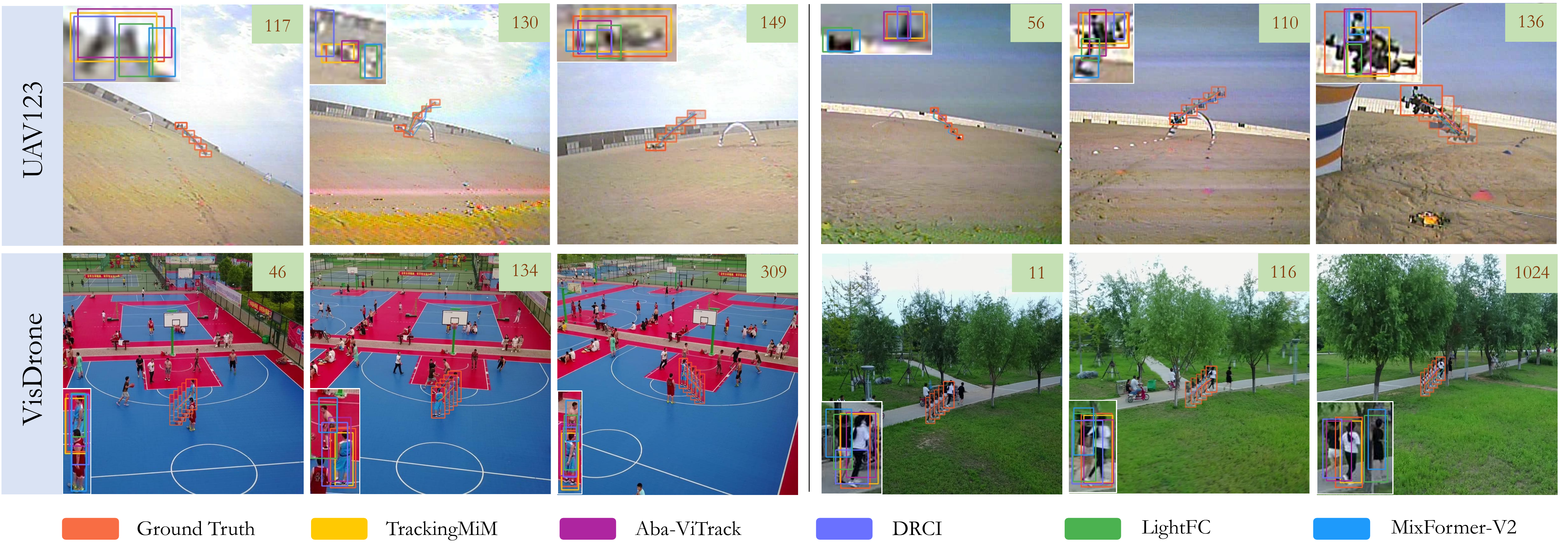}
    \caption{Qualitative results on 4 video sequences from UAV123 and VisDrone2018 (\ie \textit{uav4}, \textit{uav5}, \textit{uav0000086\_00870\_s}, and \textit{uav0000024\_00000\_s}).}

    \label{fig:visualization_other}
\end{figure*}

\subsection{Quantitative Results}
\label{quantitative}

In this section, we conduct a comprehensive evaluation of \ourmethod{} against existing lightweight trackers on multiple validation datasets. The quantitative results are summarized in Tab.~\ref{tab:main}. We compare the precision and success rate and also average FPS of CPU and GPU.

Our \ourmethod{} consistently outperforms all existing trackers across all benchmarks in terms of average precision (Prec.) and success rate (Succ.). Among DCF-based methods, RACF \cite{li2022learning} achieves $73.8\%$ Prec. and $52.8\%$ Succ. HCAT \cite{chen2022efficient} and UDAT \cite{ye2022unsupervised} achieve the highest Succ. of $62.1\%$ and highest Prec. of $80.7\%$, respectively. Among ViT-based methods, Aba-ViTrack \cite{li2023adaptive} performs best with $85.4\%$ Prec. and $64.9\%$ Succ. Our method, built on the Mamba architecture, further improves performance to 86.3\% Prec. and 66.1\% Succ.

It is noteworthy that our Mamba-based trackers achieve real-time performance at over $95$ FPS on a single CPU, outperforming the fastest ViT-based (BDTrack~\cite{wu2024learningmotionblurrobust}, $63.9$ FPS) and CNN-based (DRCI~\cite{zeng2023towards}, $64.1$ FPS) trackers. Compared to DCF-based methods, our approach is faster than most, with only KCF~\cite{henriques2015high} ($615.0$ FPS) and ECO\_HC~\cite{danelljan2017eco} ($183.8$ FPS) running ahead. On GPU, our method reaches $268.3$ FPS, comparable to DRCI ($290.6$ FPS) and BDTrack ($287.2$ FPS).

We present Precision and Success Rate curves in Fig.~\ref{fig:result_plot} to evaluate tracker performance. The precision curve measures center location error, while the success curve reports the proportion of frames with Intersection over Union (IoU) exceeding thresholds from $0$ to $1$, using the Area Under Curve (AUC) for comparison. Our \ourmethod{} achieves an average Precision AUC of $0.850$ and Success AUC of $0.666$. This surpasses the second-best tracker, Aba-ViTrack~\cite{li2023adaptive}, which reaches $0.836$ and $0.644$, with relative improvements of $1.67\%$ and $3.42\%$, respectively.

\subsection{Qualitative Results}
\label{qualitative}

To provide an intuitive understanding of tracker performance, we present qualitative comparisons across representative scenarios from multiple benchmarks. We visualize the predictions of the top five trackers, overlaid with ground-truth bounding boxes in distinct colors. Fig.~\ref{fig:visualization_dtd} shows examples from the DTB70 dataset, while Fig.~\ref{fig:visualization_other} presents results from VisDrone and UAV123. Each frame is randomly selected, with small white bounding boxes indicating predicted locations and colored trajectories illustrating the tracking paths over time. These visualizations show that \ourmethod{} consistently produces bounding boxes closest to the ground truth, even under challenging conditions such as occlusion, scale variation, multiple similar objects, and small target size.

\begin{figure*}
    \centering
    \includegraphics[width=1\linewidth]{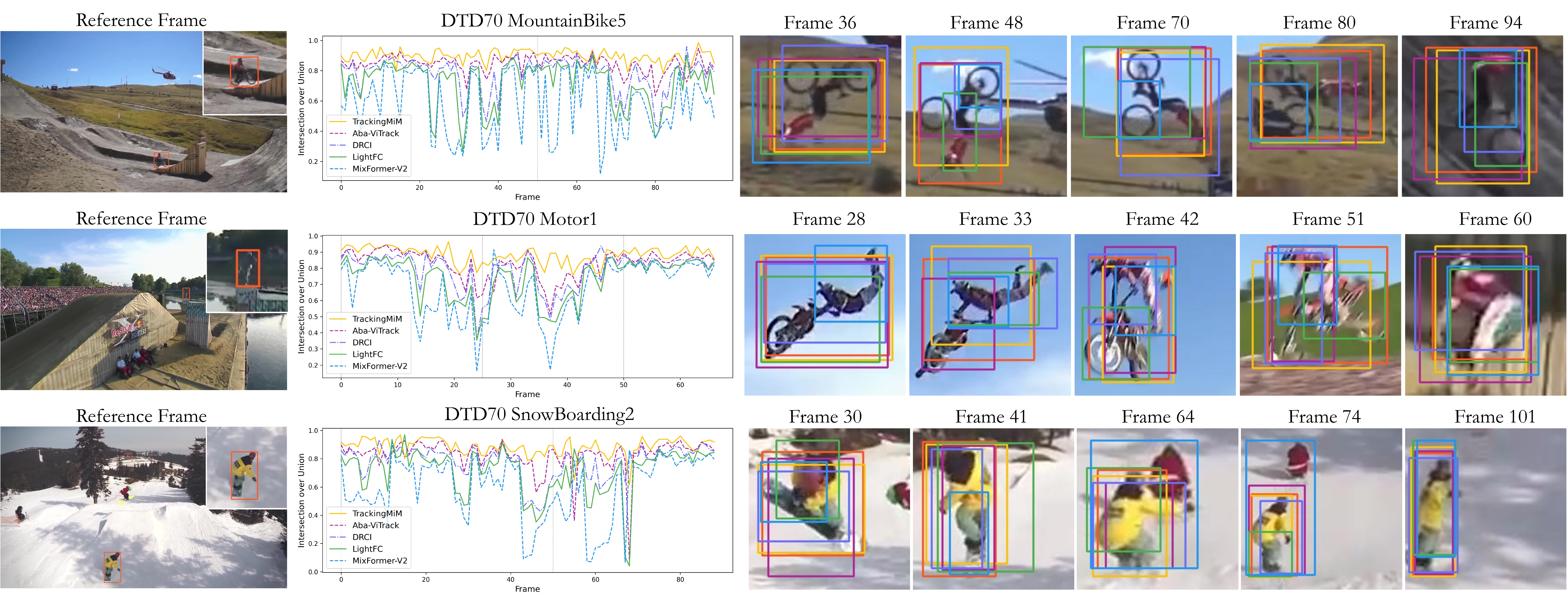}
\caption{The center plot shows IoU trends over time for videos from DTB70, with each tracker represented by a distinct color. The leftmost column displays the reference frame at $t{=}0$, used as the initial input with the target object. The five columns on the right show representative frames illustrating challenging scenarios, such as changes in object posture and the presence of similar-looking distractors.}
    \label{fig:DTD70_trend}
\end{figure*}

Fig.~\ref{fig:DTD70_trend} presents the IoU curves over time for three video examples from the DTB70 dataset. Besides  the plot, we show frame-wise tracking predictions and ground truth for visual comparison. In challenging scenarios—such as the presence of distractor objects (\eg additional people) or significant appearance changes (\eg  pose variation while driving), most trackers lose the target. In contrast, \ourmethod{} maintains accurate localization, benefiting from its temporal modeling and tracking attention. As shown in the trend plot, the orange line (ours) consistently achieves high performance throughout the sequence.

\newcommand{\p}[1]{   {\scriptsize\textcolor[RGB]{12,180,22}{(+#1)}}     }
\newcommand{\m}[1]{   {\scriptsize\textcolor[RGB]{180,12,22}{(-#1)}}     }

\newcommand{\greenm}[1]{   {\scriptsize\textcolor[RGB]{12,180,22}{(-#1)}}     }

\begin{table*}[t]
\scriptsize
\centering
\setlength\tabcolsep{5pt} 
\caption{
Ablation study on the plug-and-play integration of Tracking Attention into six high-performance trackers from CNN- and ViT-based methods. Numbers in parentheses indicate performance gains relative to the original models without Tracking Attention. All methods show improvements of over +1.0 in precision and +0.9 in success rate.
} 
\label{tab:tracking}

\resizebox{\textwidth}{!}{
\begin{tabular}{c|c|cc|cc|cc|cc|cc|cc|cc}
\toprule[1pt] 
\multicolumn{1}{c|}{}  & & \multicolumn{2}{c|}{DTB70}& \multicolumn{2}{c|}{UAVDT}& \multicolumn{2}{c|}{VisDrone} & \multicolumn{2}{c|}{UAV123}& \multicolumn{2}{c|}{UAV123@10fps}& \multicolumn{2}{c|}{Avg.}&\multicolumn{2}{c}{Avg. FPS} \\

\multicolumn{1}{c|}{\multirow{-2}{*}{Method}} & \multirow{-2}{*}{Source} & Prec.& Succ.& Prec.& Succ.& Prec.& Succ.& Prec.& Succ.& Prec.& Succ.& Prec.& Succ.& GPU & CPU  \\  \midrule

  TCTrack\cite{cao2022tctrack} & CVPR 22& 83.0 & 66.3 & 75.9 & 56.7 & 83.8 & 63.8 & 83.9 & 62.8 & 83.6 & 63.0 & 82.0\p{2.0} & 62.5\p{1.5} & 147.0\p{3.7}  & - \\
   UDAT\cite{ye2022unsupervised} & CVPR'22& {85.2} & {66.6} & 85.2 & 60.4 & 83.0 & {65.3} & 79.3 & 60.2 & 83.3 & 63.1 & 83.2\p{2.5} & 63.1\p{1.7} & 36.2\p{1.1} & - \\
  
  DRCI\cite{zeng2023towards} & ICME'23& 83.2 & 63.8 & {86.0} & {61.9} & {87.9} & 64.9 & 78.0 & 62.7 & 75.4 & 56.5 & 82.1\p{2.1} & 61.9\p{1.6} & 306.9\p{16.3} & 66.2\p{2.1} \\ \midrule

   Aba-ViTrack\cite{li2023adaptive}& ICCV'23&  {86.6} & {67.4} & {84.3} & {61.2} & {87.1} & {65.9} & {87.8} & {67.7} & {86.2} & {66.8} & {86.4}\p{1.0} & {65.8}\p{0.9} & 189.9\p{4.9} & 54.8\p{2.1}  \\
   
   LiteTrack~\cite{wei2023litetrack}  & arXiv'23  & {86.0} & {66.9} & 83.8 & 61.3 & 83.4 & 62.8 & {86.2} & {68.4} & {86.1} & {66.4} & {85.1}\p{2.5} & {65.2}\p{1.6} & 152.0\p{4.4}  & - \\
   
  LightFC\cite{Li2024LightweightFS} & KBS'24 & 84.2 & 64.7 & {86.3} & {61.4} & {83.4} & {66.4} & {89.9} & {66.8} & {83.7} & {64.9} & {85.5}\p{1.6} & {64.8}\p{1.5} & 161.3\p{8.3} & - \\

\bottomrule[1pt]


\end{tabular}}
\end{table*}


\subsection{Ablation Study}
\label{ablation}

We conduct ablation studies in Tab.~\ref{tab:main} to evaluate the impact of tracking attention and temporal modeling in MiM blocks. Both components contribute independently and jointly to performance gains. Adding temporal scanning improves precision and success by $+2.0$ and $+1.5$, respectively. Incorporating retrieval-based tracking attention yields gains of $+2.9$ and $+1.6$. Combining both achieves the highest improvement, with $+4.0$ in precision and $+2.1$ in success.

We further validate that the proposed tracking mechanism is plug-and-play. It is integrated into 6 state-of-the-art trackers, with results summarized in Tab.~\ref{tab:tracking}. Adding the tracking mechanism consistently improves performance, with at least $+1.0$ in precision and $+0.9$ in success rate. On average, it yields gains of $+1.95$ (Prec.) and $+1.47$ (Succ.), with only a modest runtime cost (approximately a 3\% reduction in FPS).


We research in the Effectiveness of MiM block design parameters (first row) and retrieval-based attention machinism (second row) with Different Components or  in Tab.~\ref{tab:ablation}. 

We study the impact of MiM block design parameters, including patch size, layer depth, and temporal window size. The results highlight a clear speed–accuracy trade-off: smaller patch sizes, deeper networks, and larger temporal windows improve performance but significantly reduce FPS. We adopt balanced configurations that offer strong tracking accuracy while maintaining real-time speed. As shown in Tab.~\ref{tab:abl_patch}, reducing the patch size from $36$ to $24$ improves precision and success by $+0.2$ and $+0.5$, respectively, but lowers FPS by $66.5$. In Tab.~\ref{tab:abl_depth}, increasing the layer depth from $24$ to $36$ yields gains of $+0.4$ (Prec.) and $+0.7$ (Succ.), with a drop of $116.1$ FPS. Similarly, Tab.~\ref{tab:abl_frame} shows that increasing the frame window size from $8$ to $16$ improves performance by $+0.3$ (Prec.) and $+0.1$ (Succ.), while reducing FPS by $83.4$.

We further evaluate the effect of the retrieval-based tracking attention parameter $K$ in Tab.~\ref{tab:abl_track_K}. When $K$ is too small ($K{<}7$), it limits the information available for retrieval. Conversely, setting $K$ too large introduces noisy features, leading to performance degradation. In all cases, changing $K$ has minimal impact on FPS. Based on this trade-off, we select $K{=}7$ as the optimal setting.

Tab.~\ref{tab:abl_track_memory} compares different feature fusion strategies. Simple mean fusion suffers from noise accumulation, achieving only $77.2$ (Prec.) and $60.5$ (Succ.). Cosine-decay fusion improves performance to $82.8$ and $62.9$, while combining it with $K$-retrieval further boosts results to $84.2$ and $64.7$, albeit with reduced FPS due to additional computation. Our final configuration—$K$-retrieval with mean fusion—achieves the best accuracy at $85.8$ (Prec.) and $65.1$ (Succ.), with only a $2.5$ FPS drop compared to the fastest baseline (simple mean).

In Tab.~\ref{tab:abl_track_injection}, we compare different strategies for injecting the fused feature into the Mamba block. Using it as an additive input is the most efficient ($281.6$ FPS) but results in the largest performance drop: $-10.3$ (Prec.) and $-4.4$ (Succ.) compared to our method. Concatenation increases computational cost without meaningful gain, yielding $-1.8$ (Prec.) and $-2.3$ (Succ.). Using the fused feature as key-value pairs in attention performs worst, with drops of $-12.4$ (Prec.) and $-5.9$ (Succ.). In contrast, our approach utilizes the fused feature as a query in cross-attention—achieves the best performance, with only a $13.3$ FPS reduction compared to the fastest baseline (simple mean fusion).

\begin{figure}
    \centering
    \includegraphics[width=1\linewidth]{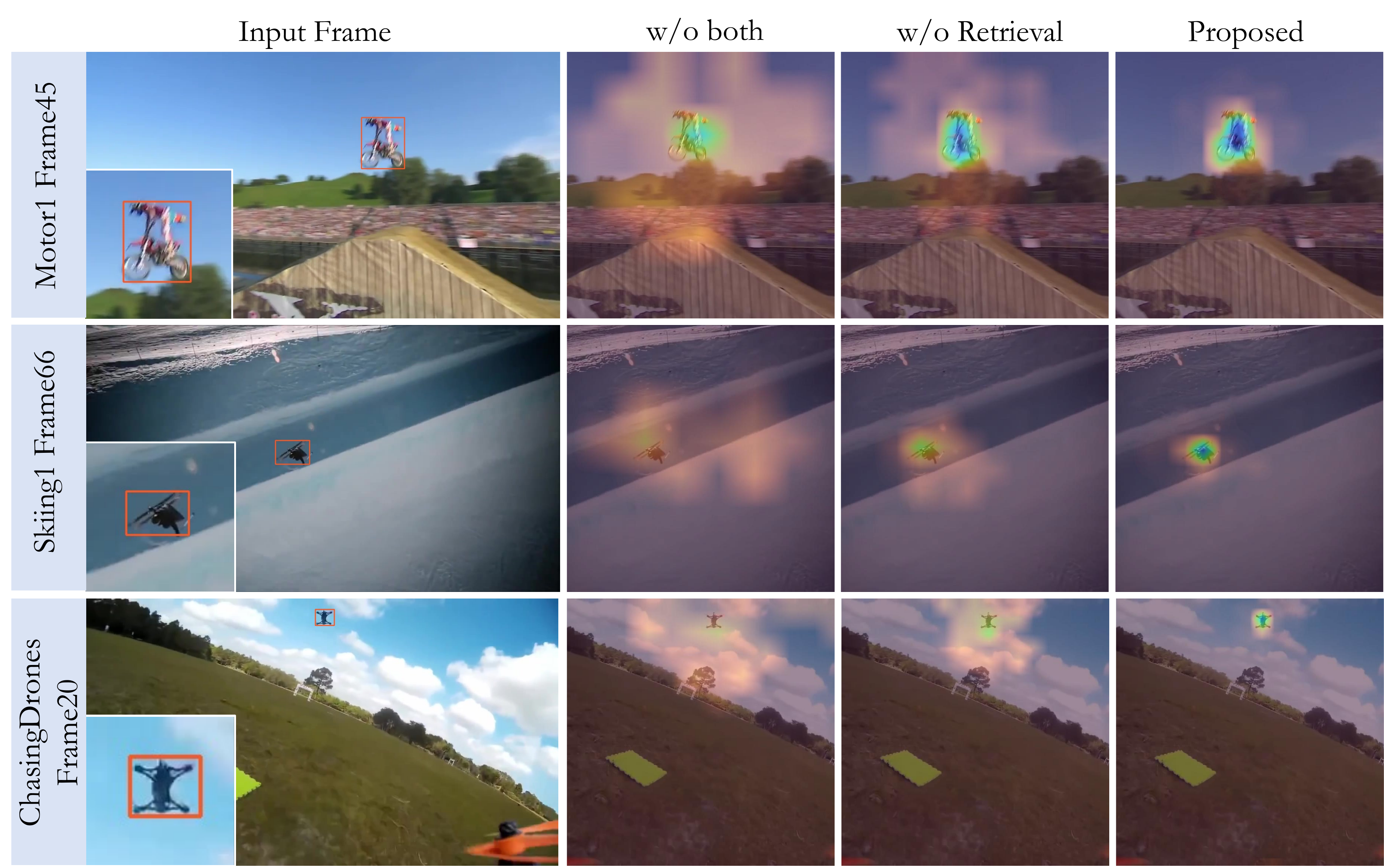}
 \caption{Class activation maps (CAM) for the ablation study. From left to right: the input frame with ground-truth bounding box, W/o Both (spatial-only Mamba block without temporal scan or retrieval), W/o Retrieval (Mamba block with spatiotemporal scanning but no tracking attention), and the Proposed method (full model with tracking attention and MiM incorporating both spatial and temporal scans).}
    \label{fig:heatmap}
\end{figure}

\subsection{Interpretable Study}
\label{interpretable}

We visualize the attention maps in Fig.~\ref{fig:heatmap} using class activation mapping (CAM) under different architectural configurations. The left-most column shows the original search images and zoomed-in target. The second column presents results from the baseline spatial-only Mamba model (w/o both) , followed by the spatial-temporal variant (w/o Retrieval), and finally our full model with object feature memory and tracking attention (Proposed).

While the spatial-only baseline roughly identifies the object, the attention maps are diffuse and uncertain, lacking clear boundaries due to the absence of temporal context. Adding temporal modeling sharpens the focus, producing more concentrated and confident maps around the target. The final configuration, with tracking attention and memory, yields the most precise localization, with attention map strongly aligned to the object.

\newcommand{\e}[1]{   {\scriptsize\textcolor[RGB]{180,180,180}{(±#1)}}     }

\newcommand{\redp}[1]{   {\scriptsize\textcolor[RGB]{180,12,22}{(+#1)}}     }

\begin{table*}[t]
\vspace{-.2em}
\centering

\caption{\textbf{Ablation studies} on key components of our Mamba-in-Mamba (MiM) architecture and retrieval-based tracking attention. We report average Precision (Prec.), Success (Succ.), and GPU FPS over five datasets. In MiM, smaller patches, deeper layers, and longer temporal windows improve accuracy at the cost of speed, highlighting a trade-off between performance and efficiency. For tracking attention, we adopt $K=7$ retrieval, apply mean aggregation over features, and use query-based attention for injection. \label{tab:ablation}
}
\subfloat[\textnormal{\footnotesize\textbf{Patch Sizes}.
Smaller patches enhance spatial resolution and accuracy but reduce speed. A patch size of $36$ achieves the best balance between performance and efficiency.} \label{tab:abl_patch}]{
\begin{minipage}{0.31\linewidth}
\begin{center}
\tablestyle{2pt}{1.05}
\resizebox{\textwidth}{!}{
\begin{tabular}{c|ccc}
Patch Size & Prec. & Succ. & FPS \\
\shline
24  & 86.0\p{0.2} & 65.6\p{0.5} & 201.8\m{66.5} \\
\rowcolor{gray!15} 
36  & 85.8\e{0.0} & 65.1\e{0.0} & 268.3\e{0.0}  \\
48  & 84.9\m{0.9}  & 64.4\m{0.7}  & 297.2\p{28.9}
\end{tabular}
}
\end{center}
\end{minipage}
}
\hspace{.25em}
\subfloat[\textnormal{\footnotesize\textbf{MiM Block Depth}. While deeper blocks improve representation quality and precision, they introduce substantial computational cost, leading to a notable drop in FPS. } \label{tab:abl_depth}]{
\begin{minipage}{0.31\linewidth}
\begin{center}
\tablestyle{2pt}{1.05}
\resizebox{\textwidth}{!}{
\begin{tabular}{c|ccc}
Layer Depth & Prec. & Succ. & FPS \\
\shline
12  & 83.1\m{2.7} & 63.4\m{1.7} & 322.7\p{54.4} \\
\rowcolor{gray!15}
24  & 85.8\e{0.0} & 65.1\e{0.0} & 268.3\e{0.0} \\
36  & 86.2\p{0.4} & 65.8\p{0.7} & 152.2\m{116.1}
\end{tabular}
}
\end{center}
\end{minipage}
}
\hspace{.25em}
\subfloat[\textnormal{\footnotesize\textbf{Frame Window}. While increasing the temporal window provides additional tracking context, performance saturates beyond 8, offering minimal improvement at the cost of significantly lower FPS.} \label{tab:abl_frame}]{
\begin{minipage}{0.31\linewidth}
\begin{center}
\tablestyle{2pt}{1.05}
\resizebox{\textwidth}{!}{
\begin{tabular}{c|ccc}
Window Size & Prec. & Succ. & FPS \\
\shline
4 & 84.5\m{1.3} & 64.6\m{0.5}  & 317.5\p{49.2}  \\
\rowcolor{gray!15} 8 & 85.8\e{0.0} & 65.1\e{0.0} & 268.3\e{0.0} \\
16 & 86.1\p{0.3} & 65.2\p{0.1} & 184.9\m{83.4} \\
\end{tabular}
}
\end{center}
\end{minipage}
}\\
\subfloat[\textnormal{\footnotesize\textbf{Feature Retrieval Top-$K$}. Precision improves with larger retrieval size for $K{\leq}7$, but degrades for $K{>}7$ due to inclusion of less relevant features, though minimal FPS impact. } \label{tab:abl_track_K}]{
\begin{minipage}{0.31\linewidth}
\begin{center}
\tablestyle{2pt}{1.05}
\resizebox{\textwidth}{!}{
\begin{tabular}{c|ccc}
$K$-retrieval & Prec. & Succ. & FPS \\
\shline
3 & 85.2\m{0.6} & 65.0\m{0.1} & 268.6\p{0.3} \\
5 & 85.6\m{0.2} & 65.0\m{0.1} & 268.5\p{0.2} \\
\rowcolor{gray!15} 7 & 85.8\e{0.0} & 65.1\e{0.0} & 268.3\e{0.0}  \\
9 & 85.6\m{0.2} & 64.7\m{0.4} & 267.8\m{0.5} \\
\end{tabular}
}
\end{center}
\end{minipage}
}
\hspace{.25em}
\subfloat[\textnormal{\footnotesize\textbf{Memory Fusion}. $K$-retrieval averaging underperforms compared to cosine-decayed and simple memory mean aggregation, which better achieve accuracy and speed. } \label{tab:abl_track_memory}]
{
\begin{minipage}{0.31\linewidth}
\begin{center}
\tablestyle{2pt}{1.05}
\resizebox{\textwidth}{!}{
\begin{tabular}{c|ccc}
Memory Fusion & Prec. & Succ. & FPS \\
\shline
Simply Mean    &77.2\m{8.6} & 60.5\m{4.6} & 265.8\p{2.5} \\
Cosine Decay        & 82.8\m{3.0} & 62.9\m{2.2} & 250.7\m{17.6} \\
\rowcolor{gray!15} 
$K$-retrieval Mean  & 85.8\e{0.0} & 65.1\e{0.0} & 268.3\e{0.0} \\
$K$-retrieval Decay  & 84.2\m{1.6} & 64.7\m{0.4} & 264.1\m{4.2}  \\
\end{tabular}
}
\end{center}
\end{minipage}
}
\hspace{.25em}
\subfloat[\textnormal{\footnotesize\textbf{Tracking Mechanism}. Injecting track features via query attention yields the best accuracy, while other mechanisms (\eg concatenate) compromise precision or speed.} \label{tab:abl_track_injection}]{
\begin{minipage}{0.31\linewidth}
\begin{center}
\tablestyle{2pt}{1.05}
\resizebox{\textwidth}{!}{
\begin{tabular}{c|ccc}
Feature Injection & Prec. & Succ. & FPS \\
\shline
Additive & 75.5\m{10.3} & 60.7\m{4.4}  & 281.6\p{13.3} \\
Concatenate & 81.0\m{1.8} & 62.8\m{2.3}  & 232.4\m{35.7}  \\
\rowcolor{gray!15} 
Q Attention & 85.8\e{0.0} & 65.1\e{0.0} & 268.3\e{0.0} \\
K-V Attention & 73.4\m{12.4} & 59.2\m{5.9} & 265.2\m{3.1}  
\end{tabular}
}
\end{center}
\end{minipage}
}
\vspace{-1.0em}
\end{table*}

\section{Conclusions}
In this work, we explore an efficient Mamba-based architecture for UAV object tracking and introduce a retrieval-augmented tracking (RAT)  mechanism to enhance re-identification during tracking. Specifically, we propose the Mamba-in-Mamba (MiM) block, which performs spatial and temporal bi-directional scanning, combined with a retrieval-based attention module. This module selects the top-$K$ features and applies mean fusion to construct the query for cross-attention, guiding accurate object localization. 
Extensive experiments on five challenging UAV benchmarks demonstrate the effectiveness of our approach. Our proposed tracker achieves state-of-the-art performance with $86.3$ precision and $66.1$ success, while maintaining high efficiency at $268.3$ FPS. Additionally, we show that our retrieval-based tracking attention can serve as a plug-and-play module, improving six existing CNN- and ViT-based trackers by at least $+1.0$ (Prec.) and $+0.9$ (Succ.) with only a $\sim3\%$ drop in FPS. We hope this Mamba-based framework inspires further research into efficient tracking using temporal and retrieval cues, especially for aerial and even general object tracking scenarios.



\section*{Acknowledgments}


The work was supported in part by the Major Project of Anonymous Institute (99999999), and the Anonymous Key Laboratory of Robotics and Computer Vision (9999999999999).




\newpage

\bibliographystyle{IEEEtran}
\bibliography{reference}

\vfill

\end{document}